\documentclass{article}
\usepackage{ijcai23}

\usepackage{times}
\usepackage{soul}
\usepackage{url}
\usepackage[hidelinks]{hyperref}
\usepackage[utf8]{inputenc}
\usepackage[small]{caption}
\usepackage{graphicx}
\usepackage{amsmath}
\usepackage{amsthm}
\usepackage{booktabs}
\usepackage{algorithm}
\usepackage{algorithmic}
\usepackage[switch]{lineno}

\usepackage{todonotes}

\usepackage{multirow}
\usepackage{multicol}

\title{Robustness and Generalization Performance of Deep Learning Models on Cyber-Physical Systems: A Comparative Study}

\author{
Alexander Windmann
\and
Henrik Steude\And
Oliver Niggemann
\affiliations
Helmut Schmidt University, Hamburg, Germany
\emails
\{alexander.windmann, henrik.steude, oliver.niggemann\}@hsu-hh.de
}

\begin{document}

\maketitle

\begin{abstract} 
Deep learning (DL) models have seen increased attention for time series forecasting, yet the application on cyber-physical systems (CPS) is hindered by the lacking robustness of these methods. 
Thus, this study evaluates the robustness and generalization performance of DL architectures on multivariate time series data from CPS.
Our investigation focuses on the models' ability to handle a range of perturbations, such as sensor faults and noise, and assesses their impact on overall performance. 
Furthermore, we test the generalization and transfer learning capabilities of these models by exposing them to out-of-distribution (OOD) samples. 
These include deviations from standard system operations, while the core dynamics of the underlying physical system are preserved.
Additionally, we test how well the models respond to several data augmentation techniques, including added noise and time warping.
Our experimental framework utilizes a simulated three-tank system, proposed as a novel benchmark for evaluating the robustness and generalization performance of DL algorithms in CPS data contexts. 
The findings reveal that certain DL model architectures and training techniques exhibit superior effectiveness in handling OOD samples and various perturbations. 
These insights have significant implications for the development of DL models that deliver reliable and robust performance in real-world CPS applications.
\end{abstract}

\section{Introduction}
\label{intro}

Cyber-physical systems (CPS) are complex systems that integrate physical and computational components, such as autonomous vehicles, power grids, and industrial control systems
\cite{niggemannMachineLearningCyberPhysical2023}. 
These systems must operate reliably and accurately in dynamic and unpredictable environments, where the consequences of failure can be severe. 
Deep learning (DL) has shown great potential for improving the performance and robustness of CPS by enabling systems to learn from data and adapt to changing conditions \cite{khanRobustnessAIbasedPrognostic2021}. 
However, the deployment of DL algorithms in CPS is not without challenges, as they must operate efficiently in real-time, handle large amounts of data, and provide accurate predictions despite noisy or adversarial input and varying environmental conditions. 

One of the critical challenges in deploying DL algorithms in CPS is the ability to generalize to new and unseen scenarios \cite{bansalSystematicGeneralizationNeural2021}. 
In real-world applications, the distribution of the data may change slowly over time or suddenly if unexpected events occur, yet the the models must be able to adapt to these changes and maintain their accuracy and reliability. 
Furthermore, in contrast to other application areas of DL like computer vision or natural language processing, DL models in CPS generally have to learn from a domain-specific dataset which only contains sensor data from one system \cite{zimmeringGeneratingArtificialSensor2021}. 
Once this system changes, e.g. because a sensor is added, usually the whole DL model has to be retrained from scratch. 
What is needed is a model that generalizes from the initial dataset, i.e. it can transfer learned knowledge about the physical systems to an altered system or a different application.

In this paper, we focus on evaluating the systematic generalization performance of several DL algorithms on a simulation of a three-tank system which follows a sequence of process phases, changing the in- and outflow of the tanks in a predictable manner. 
This system presents several challenges for DL algorithms, including nonlinear dynamics, ad-hoc process phase changes, long-term trends and noisy sensor measurements. 
We test the algorithms on new scenarios that they have not seen during training to assess their ability to capture system dynamics, learn from limited data and handle out-of-distribution (OOD) samples.

The research questions (RQ) to be discussed in this paper can be summarized as follows:
\begin{enumerate}
    \item What deep learning model architectures demonstrate robustness to perturbations in CPS, specifically noise, variation on timings and faulty sensors, and how much do these scenarios impact the models' performance?
    \item How effectively can these DL model architectures generalize to unfamiliar system behaviors, which may include systems operating at different scales or presenting new combinations of process phases, all while ensuring the fundamental system dynamics are maintained?
    \item Which DL model architectures exhibit the highest capability to transfer system-specific learned knowledge to related tasks following fine-tuning?
    \item To what extent do data augmentation techniques during training improve the robustness and generalization capabilities of different DL model architectures?
\end{enumerate}

We hope that this study can help to uncover the key design principles and modeling approaches that can improve the robustness and reliability of DL models in CPS, especially when dealing with OOD samples.
Furthermore, by making our code and the simulation environment available openly, we provide a benchmark for researchers of robust DL models on CPS.
\footnote{Code: https://github.com/alexwndm/robust-ml-for-cps}

The structure of this paper is as follows: 
Section 2 surveys relevant work, underscoring the research gaps this study aims to bridge. Section 3 describes our simulation environment, the selected models, and our comprehensive experimental setup. 
Section 4 presents the outcomes of our experiments alongside a detailed analysis. 
In Section 5, we summarize our findings, draw conclusions, and propose potential avenues for future research.

\section{Related Work}
\label{related}
Due to the rising complexity of CPS, there is an increasing demand on using DL in industrial applications, such as anomaly detection \cite{luoDeepLearningbasedAnomaly2021}, predictive maintenance \cite{debeaulieuUnsupervisedRemainingUseful2022} and planning \cite{kocherResearchAgendaAI2022}. 
As CPS share many similarities with dynamical systems, most DL models that are in use focus on processing time series data. 
Such models include recurrent networks like LSTM networks \cite{hochreiterLongShortTermMemory1997,chenDataDrivenApproachAssessing2022} and convolutional models, such as Temporal Convolutional Networks (TCNs) \cite{leaTemporalConvolutionalNetworks2017,mengTimeConvolutionalNetwork2020}. 
Recently, the Transformer architectures have also been gaining traction in modeling physical systems \cite{vaswaniAttentionAllYou2017,genevaTransformersModelingPhysical2022}.

Robustness is a fundamental aspect of DL methods, as it directly impacts the reliability and safety of applications\cite{bengioDeepLearningAI2021}.
Despite their effectiveness in narrow tasks, DL models often exhibit limited performance when faced with small perturbations in the input data, such as adversarial attacks \cite{tsiprasRobustnessMayBe2019,buzhinskyMetricsMethodsRobustness2021}.
This limitation presents a significant challenge in CPS, where robustness is paramount for the safe deployment of DL methods \cite{khanRobustnessAIbasedPrognostic2021}.
Particularly, CPS applications, given their diverse sensory input, are also prone to adversarial attacks \cite{liAdversarialAttacksDefenses2020}. 
Despite this, a comprehensive study on the impact of various perturbations in CPS on DL architectures is yet to be conducted.

OOD generalization is another crucial aspect in DL. 
Most models rely on the assumption of independent and identically distributed data and struggle to generalize to new, previously unseen data \cite{scholkopfCausalRepresentationLearning2021}.
When the distribution of input data deviates significantly from the training data, the performance of the models can deteriorate.
For dynamical systems, learning the underlying causal variables is one approach \cite{lippeICITRISCausalRepresentation2022,lippeCITRISCausalIdentifiability2022}, which has proven successful in CPS data applications such as prognostics and health management \cite{koutroulisConstructingRobustHealth2022}. 
Another research direction explores the learning of independent mechanisms from data \cite{goyalRecurrentIndependentMechanisms2021,goyalCoordinationNeuralModules2022}, often resulting in modular model architectures that exhibit improved OOD predictions on CPS data \cite{bansalSystematicGeneralizationNeural2021}.
However, a comprehensive empirical study of the generalization performance of DL models for CPS remains an open area for exploration.

Data augmentation is one of several learning techniques that has shown to improve robustness of DL models \cite{shortenSurveyImageData2019}. 
Beyond computer vision, it has found success in areas such as time series analysis \cite{wenTimeSeriesData2021}, Natural Language Processing (NLP) \cite{bayerSurveyDataAugmentation2022}, and graph machine learning \cite{zhaoGraphDataAugmentation2023}. 
Recent years have seen the application of data augmentation in CPS as well \cite{yinLeveragingDataAugmentation2021}.
There is, however, still a lack of works that systematically investigate the impact of data augmentation on CPS applications.

\section{Methology}
\label{methology}
\begin{figure}
  \centering
  \includegraphics[width=\columnwidth]{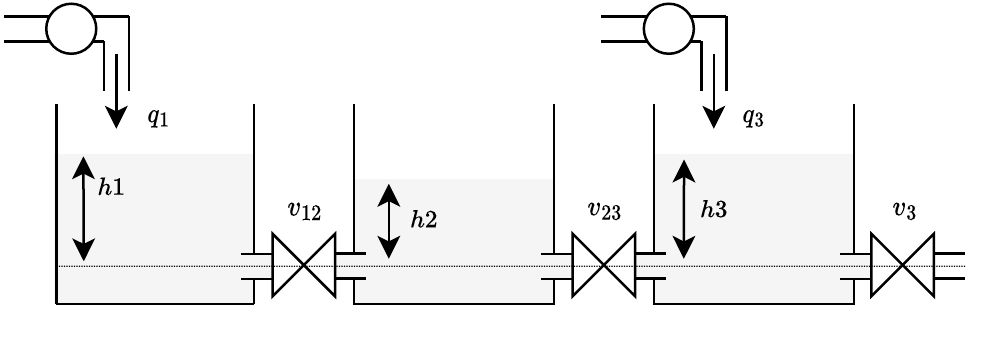}
  \caption{This figure schematically shows the three-tank system used in the experiments. It includes the unobservable inflow $q_1$ and $q_3$, the valve coefficients $v_{12}, v_{23}$ and $v_{3}$ as well as the observable water levels $h_1, h_2$ and $h_3$.}
  \label{fig:three-tank}
\end{figure}

In this study, we simulate a three-tank system, as depicted in Figure \ref{fig:three-tank}, to evaluate the robustness and generalization performance of various DL architectures on CPS.
Our primary focus is on forecasting, a task that necessitates a deep understanding of system dynamics.
Forecasting is not only analogous to various CPS-related tasks, such as industrial process control \cite{downsPlantwideIndustrialProcess1993} and prognostics and health management \cite{leePrognosticsHealthManagement2014}, but can also serve as a self-supervised learning pre-training task that could benefit other applications \cite{zerveasTransformerbasedFrameworkMultivariate2021}.

The transparent system dynamics offered by the three-tank simulation provide full control over system behavior, enabling qualitative evaluation through minor adjustments.
Furthermore, a simulation allows to evaluate the models' generalization performance systematically.
In general, evaluating the generalization performance of forecasting models is challenging, since predicting arbitrary time series is infeasible. 
However, simulations of CPS like the three-tank system share a common foundation in the underlying physical and dynamical system. 
This enables the generation of realistic OOD samples that serve as reasonable targets for testing generalization performance. 
Moreover, our simulation can be readily adapted to more complex scenarios, such as a four-tank system, thereby extending its applicability in studying DL model performance across a diverse range of CPS tasks. 
Openly accessible data from real-life CPS lacks this flexibility, as the dataset is typically fixed and intricate to the extent that the impacts of interventions cannot be reliably predicted during experiment design. 
Therefore, our simulation-based approach provides a more controlled and adaptable framework for understanding and enhancing the performance of DL models in CPS applications.

\subsection{Simulation Environment}
\label{environment}

To generate the training dataset, we employ a systematic approach, filling and mixing the tanks in a controlled and predictable manner. 
The dataset is composed of subsequential process phases, each with a duration of 50 time steps. 
During each process phase, control variables, such as inflow rates and valve coefficients, are held constant, allowing the system's dynamics to evolve according to the predefined underlying equations:

\begin{equation} \label{eq:three-tank}
\begin{aligned} 
\dot{h}_1&= q_1-k_{v_{12}} \operatorname{sign}\left(h_1-h_2\right) \sqrt{\left|h_1-h_2\right|} \\
\dot{h}_2 & = k_{v_{12}}\operatorname{sign}\left(h_1-h_2\right) \sqrt{\left|h_1-h_2\right|} \\
&\quad -k_{v_{23}}\operatorname{sign}\left(h_2-h_3\right) \sqrt{\left|h_2-h_3\right|} \\
\dot{h}_3&=   q_3+k_{v_{23}}\operatorname{sign}\left(h_2-h_3\right) \sqrt{\left|h_2-h_3\right|}-k_{v_3} \sqrt{h_3}
\end{aligned}
\end{equation}
where $h_1(t)$ $h_2(t)$ and $h_3(t)$ describe the water levels of the three tanks at the time step $t$, $q_1$ and $q_3$ describe the inflow and $k_{v_{12}}$, $k_{v_{23}}$ and $k_{v_{3}}$ describe the valve coefficients as depicted in Figure \ref{fig:three-tank}.

\begin{figure}
  \centering
  \includegraphics[width=\columnwidth]{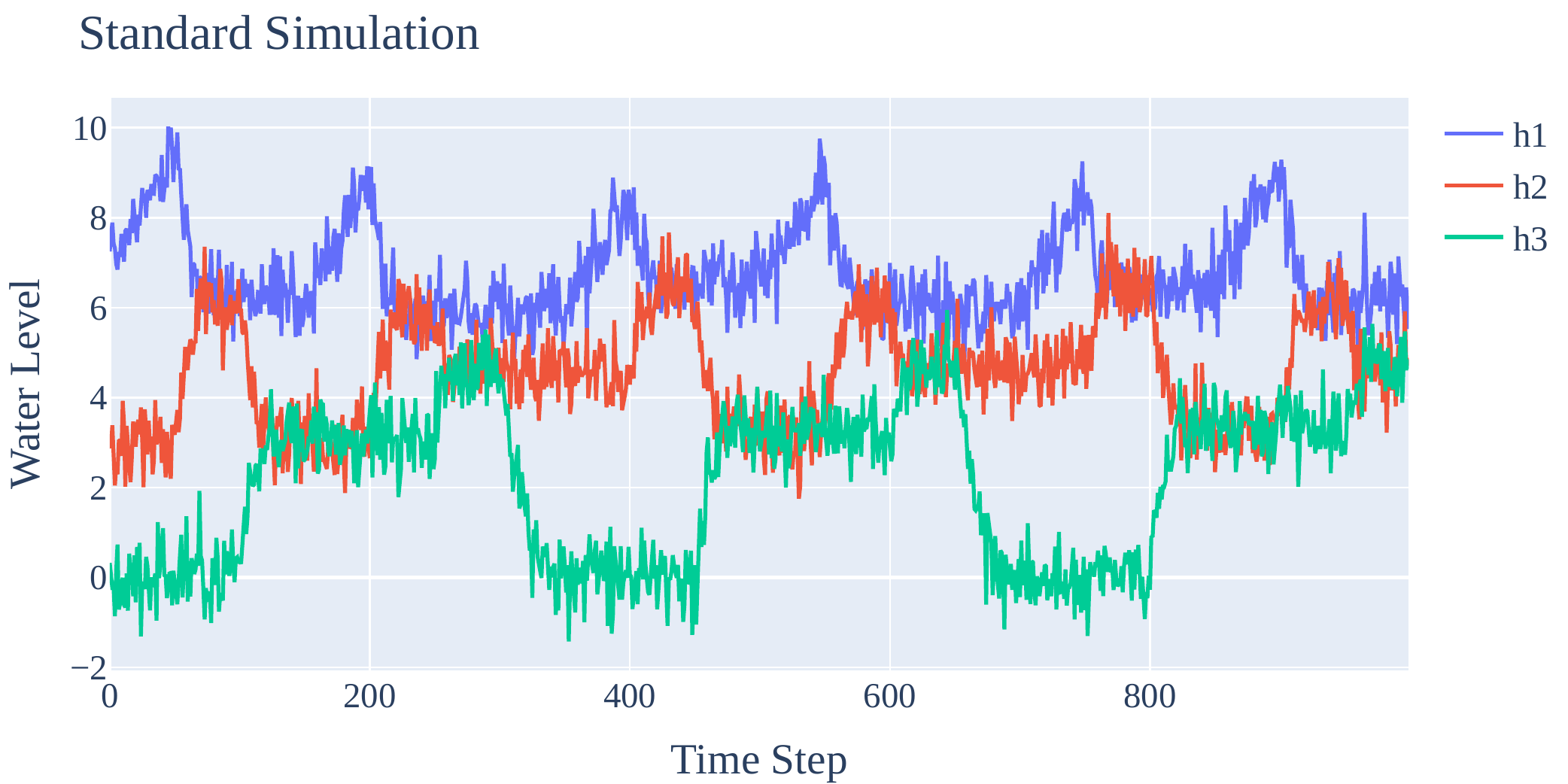}
  \caption{This figure shows a section from the 'standard' scenario dataset of the three-tank simulation. A sample consists of 250 time steps, the forecasting target is 50 time steps long.}
  \label{fig:sim_std}
\end{figure}

To create a closer resemblance to real-world CPS, we simulate production by progressing through a defined sequence of process phases: initially filling Tank 1, followed by mixing Tanks 1 and 2, then mixing Tanks 2 and 3, refilling Tank 1, repeating the two mixing stages, and ultimately emptying Tank 3.
To emulate a more realistic inflow pattern, we introduce a long-term cosine trend, representing the ramp-up and shutdown of inflow rates. 
Additionally, Gaussian noise is incorporated into sensor values to further increase the simulation's realism.
Detailed information regarding the simulation and dataset generation can be found in the associated repository, and a representative sample is illustrated in Figure \ref{fig:sim_std}. 

\subsection{Models Under Test}
\label{models}

In this study, we evaluate the robustness and generalization performance of various deep learning models using a three-tank system as the test environment. 
Guided by foundational architectural principles widely adopted in the DL research community – recurrence, convolutions, and attention mechanisms – we select a prominent model from each category. 
Furthermore, we experiment with different model variations within each category. 
The models under investigation include:

\begin{enumerate}

\item Multilayer Perceptron (MLP): As a baseline, we employ an MLP with up to three hidden layers, utilizing the Rectified Linear Unit (ReLU) activation function between layers.

\item Gated Recurrent Unit (GRU) \cite{choLearningPhraseRepresentations2014}: To evaluate recurrent architectures, we implement a GRU network. 
We adopt two forecasting approaches: (i) using a linear layer on the final hidden state to predict the entire time series directly, and (ii) employing an auto-regressive method, denoted as GRU-AR, to predict the target time series step by step.

\item Temporal Convolutional Network (TCN) \cite{leaTemporalConvolutionalNetworks2017}: To examine convolutional networks, we implement a TCN. 
We experiment with significantly reducing the dimensionality in the latent space using the autoencoder framework \cite{Bengio2013-yv} similar to TCN-AEs \cite{mengTimeConvolutionalNetwork2020} additional to a decoder for the forecasting objective.
We refer to this as the TCN-Forecasting-Autoencoder (TCN-FAE).

\item Transformer \cite{vaswaniAttentionAllYou2017}: To assess attention networks, we implement a vanilla encoder-decoder Transformer. 
Additionally, we explore embedding the time series using causal convolutions, as proposed by \cite{liEnhancingLocalityBreaking2019}.
This is denoted as Transformer-CE.


\end{enumerate}

\subsection{Experimental Setup}
\label{experiments}

The backbone of all subsequent experiments are models trained on a dataset derived from the three-tank simulation delineated in Subsection \ref{environment}. 
This time series is partitioned into training, validation, and testing intervals. 
For the training dataset, we randomly select 1000 samples from the training interval, while the validation and testing datasets each comprise 100 samples. 
Each model is fed a random 250-time-step interval as input and assigned the task of forecasting the subsequent 50 time steps of the three water levels.
Upon the completion of each epoch, we compute the Mean Squared Error (MSE) on the validation dataset. 
The model manifesting the lowest epoch validation loss across all models and epochs is chosen during hyperparameter tuning. 
All models employ the Adam optimizer \cite{DBLP:journals/corr/KingmaB14}, with the learning rate halved if the epoch validation loss does not reduce for 25 epochs \cite{Bottou2012-sv}. 
Although all models are trained for a maximum of 1000 epochs, early stopping is enforced if the validation loss ceases to decrease for 50 subsequent epochs. 
Comprehensive implementation details, including hyperparameters and training procedures, are available in the associated repository.

It should be noted that the samples from the training, validation, and testing datasets, drawn from the same simulation, bear a similar distribution. 
This contrasts with the forthcoming experiments, where models predict samples drawn from alternative simulations, termed as scenarios. 
The time series referenced in Subsection \ref{environment} will henceforth be referred to as the 'standard' scenario.

\subsubsection{Experiment 1: Robustness to Perturbations}
\label{expt:perturbations}

To address RQ 1, which pertains to the robustness of the models in the face of perturbations, we evaluate the trained models on various simulations where the time series are influenced by specific perturbations.
The scenarios are as follows:
\begin{itemize}
    \item \textbf{Scenario 1: Sensor Faults.} We introduce anomalies to the models' input data to simulate sensor faults, a common issue in real-life CPS. 
    In half of the instances, we add a point anomaly for one sensor at a random time step. 
    For the remaining instances, we simulate a dead sensor by setting its value to zero for up to 50 time steps. 
    We ensure that these anomalies are excluded from the last input steps and the target sequence, thereby evaluating each model on an accurate time series.

    \item \textbf{Scenario 2: Increased Noise.} We simulate the three-tank system with thrice the amount of Gaussian noise on all sensor values, making the forecasting task more challenging.

    \item \textbf{Scenario 3: Variable Process Phase Duration.} We manipulate the duration of different process phases, such as filling or mixing tanks, by adding noise to the 50 step process phase duration interval in the simulation.
\end{itemize}

\subsubsection{Experiment 2: OOD Generalization}
\label{expt:generalization}


To assess the generalization performance of various DL models on OOD samples and thereby answer RQ 2, we investigate their forecasting performance in different scenarios where the behavior or the settings of the underlying three-tank system change. 
The underlying causal and physical system described in Equation \eqref{eq:three-tank} serves as the foundation for each simulation, ensuring that OOD samples are challenging yet grounded in reality. 
A DL model that fully captures the underlying causal and dynamic system should exhibit improved forecasting performance in these new scenarios.

The scenarios are as follows:
\begin{itemize}
    \item \textbf{Scenario 4: Scaled Inflow and Outflow.} We increase both the inflow and valve coefficients by 50\%. 
    This results in generally higher water levels while preserving the system's behavior.
    \item \textbf{Scenario 5: Independent Process Phase Merging.} We simultaneously fill Tank 1 and empty Tank 3, combining two independent process phases.
    \item \textbf{Scenario 6: Independent Process Phase Merging with Additional Stable Process Phase.} By combining the two process phases in the previous scenario, we effectively reduce the underlying cycle from 7 to 6 process phases. 
    To assess the effect of this change, we add an additional process phase to the cycle during which the water levels remain unchanged.
    \item \textbf{Scenario 7: Dependent Process Phase Merging.} We create a novel process phase by opening both valves between the tanks simultaneously, mixing all tanks at once. 
\end{itemize}

Additionally, to answer RQ 3 we try to evaluate the transfer learning capabilities of the models by fine-tuning them on the new scenarios. 
We do this by reusing the model weights and retraining the models on a new objective: predict the next 50 time steps of the water tanks of the new OOD scenario.
We limit the maximum number of epochs the models are allowed to train to 50 and choose the model checkpoint with the best loss on the validation dataset of the OOD scenario. 
Models that have better learned the underlying concepts, such as mixing or filling of tanks, should adapt more quickly when fine-tuned to the new system.

\subsubsection{Experiment 3: Data Augmentation}
\label{expt:augmentation}

To investigate the extent to which the models described in Subsection \ref{models} can benefit from data augmentation techniques (addressing RQ 4), we select the models that have been trained under the 'standard' scenario and train them on additional augmented data. 
This augmented data was generated by applying two types of transformations on the original dataset derived from the 'standard' scenario:

\begin{enumerate}
\item \textbf{Added Noise:} We enhance the time series by adding Gaussian noise. 
This additive distortion complements the inherent noise that forms part of the simulation.
\item \textbf{Time Warping:} We introduce variability in the time series by adding random noise to the original time step intervals, thereby warping the time scale. 
The new values are obtained by interpolating the original time series values at these adjusted time steps and are subsequently mapped to the original time indices. 
The resultant time series maintains similarity to the original, albeit with more randomized timings.
\end{enumerate}

The models pre-trained under the 'standard' scenario are then fine-tuned on these synthetically augmented datasets following an approach akin to that adopted in Experiment 2.

\section{Results}
\label{results}
\subsection{Experiment 1: Robustness to Perturbations}

The results of the first experiment are outlined in Table \ref{tab:expt1}.
In the 'standard' scenario, all model architectures demonstrated the capability to accurately predict the future water level of the three tanks, with minimal variation in predictive accuracy.
Notably, the GRU model with an auto-regressive forecasting pattern generated the most precise predictions.
Interestingly, more sophisticated architectures like TCNs or Transformers did not surpass the predictive performance of the MLP.

\begin{table}[]
\centering
\begin{tabular}{lccccc}
\toprule
\textbf{Model} & \textbf{Standard} & \textbf{Scn. 1} & \textbf{Scn. 2} & \textbf{Scn. 3}\\
\midrule
MLP         & 0.32 & \textbf{0.98}  & \textbf{2.34} & \textbf{0.72}\\
GRU         & 0.33 & 2.97  & 2.48 & 0.92 \\
GRU-AR      & \textbf{0.28} & 2.05  & 2.42 & 0.89 \\
TCN         & \underline{0.38} & 5.65  & 2.77 & 1.21 \\
TCN-FAE      & 0.34 & 4.02  & 2.40 & 1.11 \\
Transformer & \underline{0.38} & \underline{15.55} & \underline{3.78} & \underline{1.28}\\
Transf.-CE  & 0.37 & 8.78  & 3.40 & 1.04\\
\bottomrule
\end{tabular}
\caption{Performance comparison of models across different scenarios (MSE). Lower values indicate better performance. The best performance for each scenario is highlighted in bold, the worst is underlined. The scenarios include: 'sensor faults' (Scn. 1), 'increased noise' (Scn. 2) and 'variable process phase duration' (Scn. 3).}
\label{tab:expt1}
\end{table}

The good performance of the MLP is further evident in the scenarios involving perturbations introduced to the three-tank simulation. 
Particularly in the presence of sensor faults, the MLP outperformed all other models, despite its performance drop.
For the simulation with faulty sensors, a clear pattern emerged across other architectures: recurrent architectures are affected by these perturbations, convolutional models performed worse and Transformer variants utterly failed to predict the time series when faced with faulty data, which partly answers RQ 1. 

\begin{figure}
  \centering
  \includegraphics[width=\columnwidth]
  {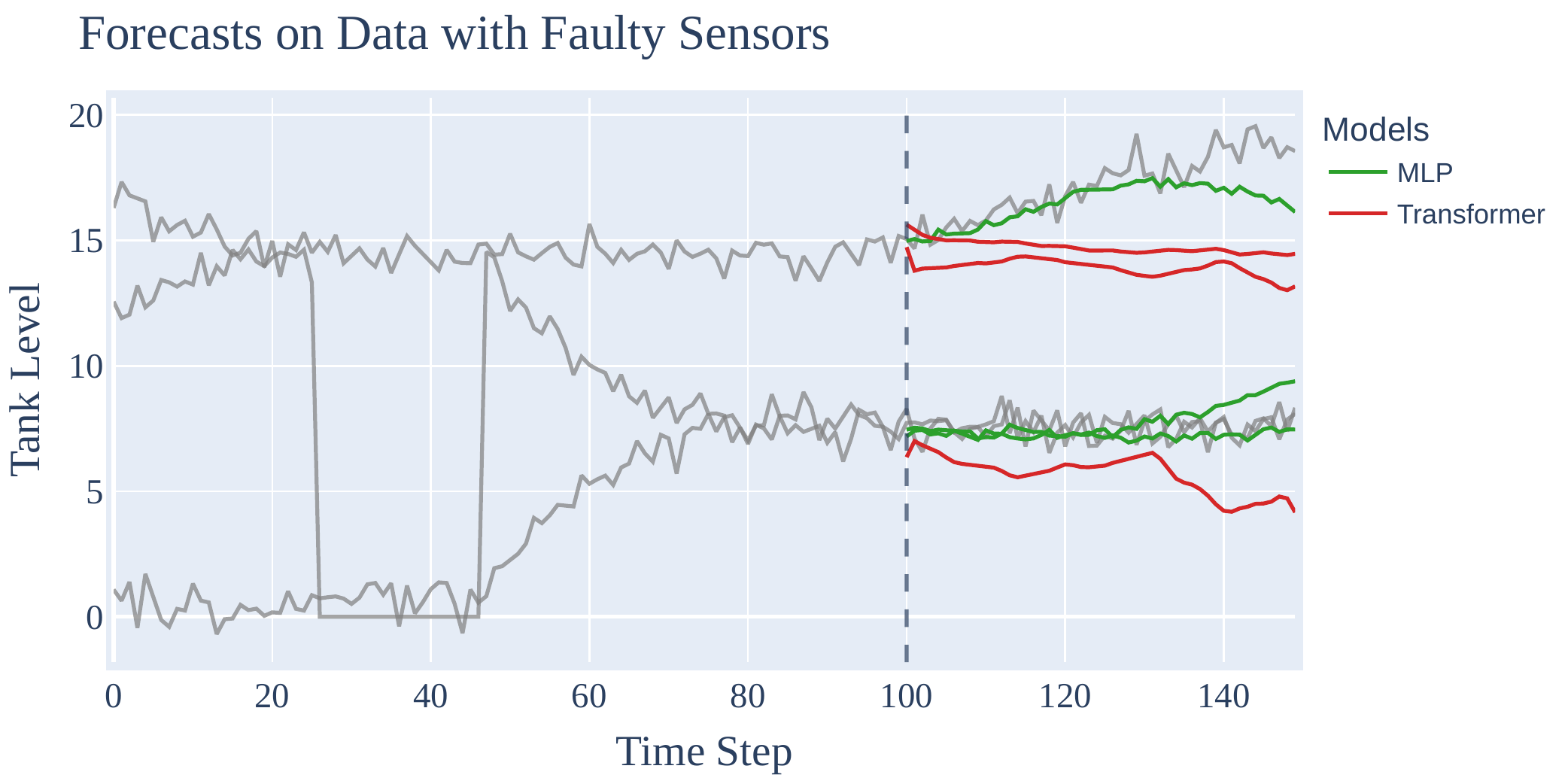}
  \caption{Sample forecast for selected models on the simulation with faulty sensors. While the MLP is mostly unaffected, the Transformer completely fails to predict the future time steps.}
  \label{fig:forecast}
\end{figure}

Figure \ref{fig:forecast} provides a graphical representation of the MLP's and Transformer's forecast when presented with a faulty sensor.
While the MLP's forecast was minimally affected, the Transformer faltered significantly in its predictions. 
It incorrectly predicted the water levels of all three tanks.
Notably, it predicted the water level of Tank 2 to start at a high level, even though the tank has just been emptied.
This underlines the model's deficiency in learning fundamental properties of the underlying physical system, as such sudden fluctuations of water levels are implausible in the standard three-tank scenario the model was trained on. 
As this performance drop has not been foreseeable during the 'standard' scenario, the experiment highlights the importance of performing extensive tests on models before they are applied in production on a real-life CPS.

In the scenarios involving heightened noise and variable process phase durations, the behavior of the models is similar to Scenario 1, albeit with less pronounced differences.
Given the increased uncertainty in Scenarios 2 and 3, a rise in MSE was anticipated as a perfect prediction of such randomness is inherently infeasible. 
Despite this, the MLP maintained its superior performance.
Transformers, on the other hand, demonstrated inferior performance under increased noise, while recurrent models outperformed other architectures, barring the MLP, under variable process phase durations, which answers RQ 1.

\begin{figure}
  \centering
  \includegraphics[width=\columnwidth]{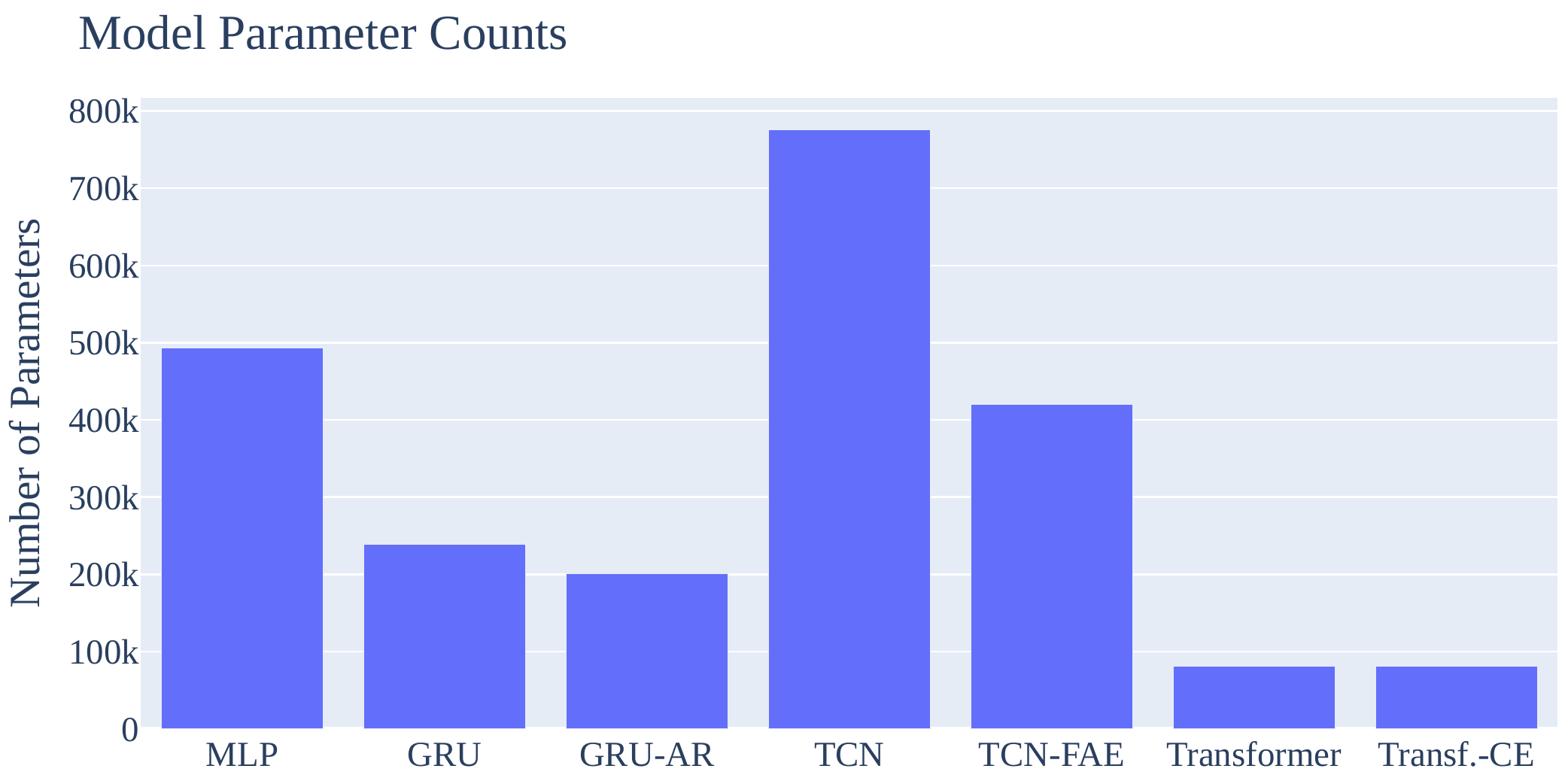}
  \caption{This figure shows the parameter count for the tested models.}
  \label{fig:param_count}
\end{figure}

In summary, an increased complexity in the model architecture can lead to impaired robustness to data perturbations, as exemplified by the MLP's performance in the tested simulations. 
However, no distinct correlation between model size and robustness was discernible across model architectures, as depicted in Figure \ref{fig:param_count}. 
Hyperparameters for all models were determined through extensive grid search, as detailed in Section \ref{experiments}. 
It was observed that larger convolutional architectures such as TCNs tended to perform better, while recurrent models were relatively smaller. 
Transformers, generally, have been noted to perform better when scaled up \cite{brownLanguageModelsAre2020}. 
However, due to hardware constraints, we were limited by the self-attention algorithm's memory and runtime of $\mathcal{O}(n^2)$ for an input sequence length of $n$. 
Given the input length of 250, even relatively small Transformer variants pushed the A100 GPU used for model testing to its limits.
This restriction, however, was universally applicable to all models, ensuring a level playing field. 
It would be intriguing to examine the performance of more recent Transformer variants with improved runtime and memory footprint when subjected to similar scenarios.

\subsection{Experiment 2: OOD Generalization}

In the subsequent experiment, we assessed the models' performance across a variety of scenarios where the behavior of the underlying three-tank system was altered. 
The findings of this investigation are detailed in Table \ref{table:expt2-summary}.
\begin{table*}[h]
\centering
\begin{tabular}{lcccccccc}
\toprule
\multirow{2}{*}{\textbf{Model}} & \multicolumn{2}{c}{\textbf{Scenario 4}} & \multicolumn{2}{c}{\textbf{Scenario 5}} & \multicolumn{2}{c}{\textbf{Scenario 6}} & \multicolumn{2}{c}{\textbf{Scenario 7}} \\
\cmidrule(lr){2-3} \cmidrule(lr){4-5} \cmidrule(lr){6-7} \cmidrule(lr){8-9}
                                & \textbf{Epoch 0} & \textbf{Epoch 50} & \textbf{Epoch 0} & \textbf{Epoch 50} & \textbf{Epoch 0} & \textbf{Epoch 50} & \textbf{Epoch 0} & \textbf{Epoch 50} \\
\midrule
MLP                             &\textbf{0.53} &\textbf{0.53}& 6.01       & 0.44            &\textbf{1.27}  & 0.43          &\textbf{6.02}  &\textbf{0.35}\\
GRU                             & 10.27        &\textbf{0.53}& 4.79       &\textbf{0.35}    & 1.28          & 0.39          &\underline{9.65}& 0.37        \\
GRU-AR                          & 14.80        &\textbf{0.53}&\textbf{2.56}&\textbf{0.35}   & 1.32          &\textbf{0.35}  & 8.12          & 0.36        \\
TCN                             & 1.52         & 1.32        & 5.18       & 0.57            &\underline{2.52}& 0.50         & 8.28          & 1.02        \\
TCN-FAE                          & 1.28         & 1.49        &\underline{6.47}&\underline{2.93}& 1.71       &\underline{1.40}& 7.21         &\underline{2.19}\\
Transformer                     & 13.06        &\underline{2.59}& 4.13    & 0.47            & 1.51          & 0.46          & 9.45          & 0.99        \\
Transf.-CE                      &\underline{19.04} & 1.66        & 4.97   & 0.44            & 1.57          & 0.40          & 8.06          & 0.79        \\
\bottomrule
\end{tabular}
\caption{Comparative performance of different models fine-tuned across various test scenarios. The table presents the MSE values obtained from models trained on the 'standard' scenario and subsequently fine-tuned for 0 or 50 epochs on the respective scenario. Lower MSE values signify better performance. For each scenario and fine-tuning duration (0 or 50 epochs), the highest performing model is denoted in bold, while the lowest performing model is underlined. The scenarios include: 'scaled inflow and outflow' (Scenario 4), 'independent process phase merging' (Scenario 5), 'independent process phase merging with an additional stable phase' (Scenario 6), and 'dependent process phase merging' (Scenario 7).}
\label{table:expt2-summary}
\end{table*}

In Scenario 4, both the inflow and the valve coefficients of the system were increased by 50\%. 
Despite the similar behavior to the learned 'standard' scenario, the sensor values were generally elevated. 
There is a clear distinction in between the MLP and the TCN variants, which yielded satisfactory predictions, and the GRU and Transformer variants, which did not meet expectations. 
Following a fine-tuning phase of 50 epochs, a significant improvement was noted in most models. 
The exceptions are the TCN-FAE, which shows a decreasing performance, and the MLP, which already demonstrated excellent performance prior to fine-tuning. 

Scenario 5 encompassed two process phases that did not directly influence each other, executed concurrently. 
This implied that models with the ability to integrate the learned behavior of two separate process phases would outperform others. 
Regrettably, none of the models predicted the new scenario accurately without the benefit of fine-tuning.
Curiously, the auto-regressive GRU variant outperforms the other models before and after fine-tuning. 
It would be interesting to further investigate the effect of this forecasting method on integrating learned behavior.

The merging of two scenarios introduced a new complexity: the system had to transition from cycling through a fixed number of process phases to one less phase. 
To explore the extent to which the models rely on this cyclic structure in their predictions, an additional 'stable' process phase was incorporated into Scenario 6. 
Hypothetically, the 'stable' process phase where the water levels in the tanks remain constant presents a new challenge for the models, as it is not part of the 'standard' scenario the models have been trained on. 
However, Table \ref{table:expt2-summary} reveals that the initial MSE values, prior to fine-tuning, were substantially lower than in Scenario 5. 
This suggests that the models have, to a significant degree, learned the underlying cyclic structure of the simulation. 
After fine-tuning, this disparity disappeared, indicating that the models are capable of swiftly adapting to changes in the cycle length.

Finally, Scenario 7 involved the merger of two dependent process phases where Tanks 1 and 2 were mixed while Tank 2 was concurrently mixed with Tank 3. 
An effective prediction of this scenario without fine-tuning would necessitate a model capable of understanding the 'mixing tanks' mechanism and applying it simultaneously from two perspectives. 
As anticipated, the models struggled more in this scenario than in Scenario 5. 
The MLP was the only model that managed to maintain a somewhat consistent performance, albeit with a high MSE.

\begin{figure}
  \centering
  \includegraphics[width=\columnwidth]{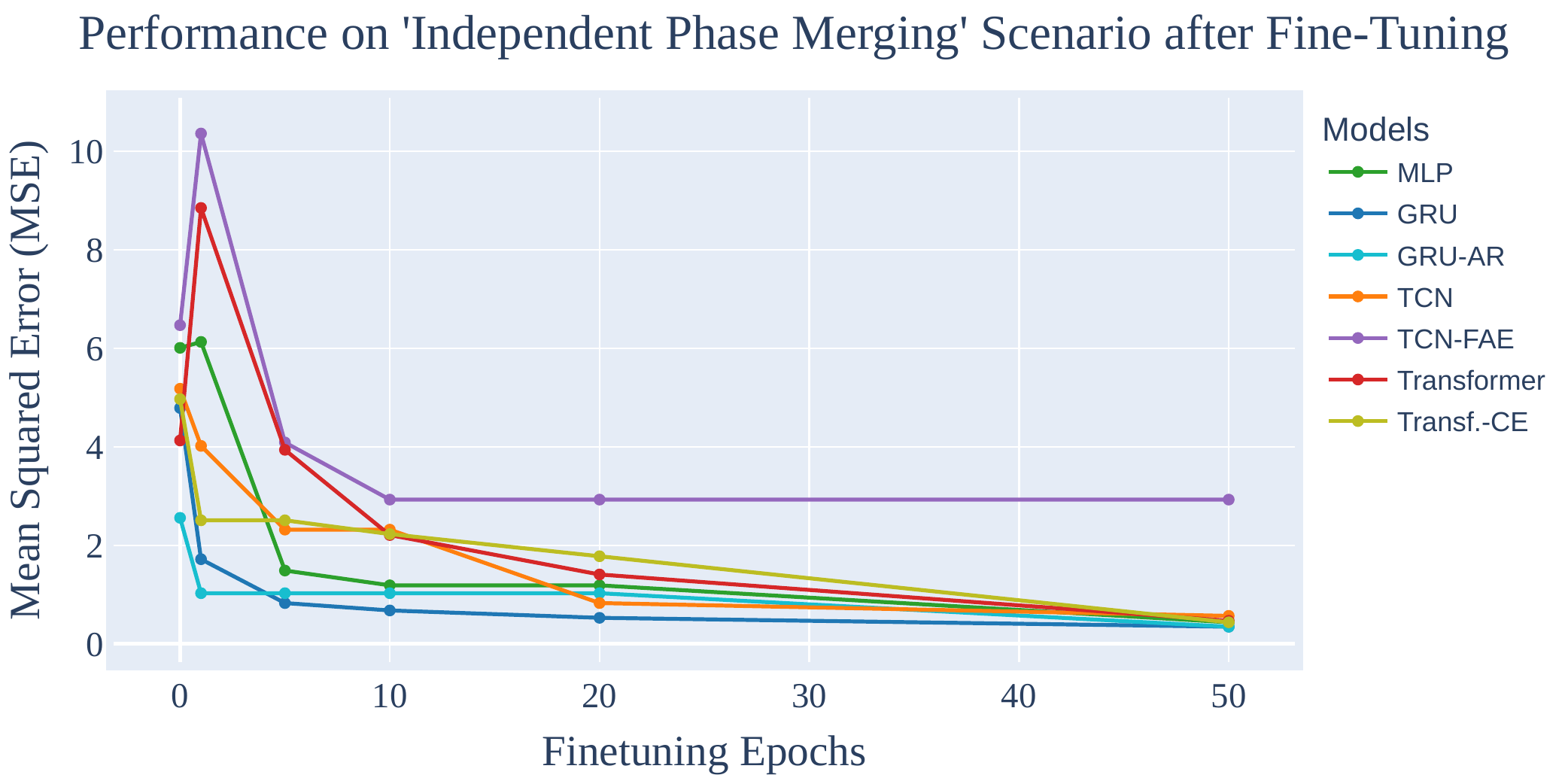}
  \caption{Performance of different models on the scenario with independent phase merging after fine-tuning. }
  \label{fig:ftune_q1+v3}
\end{figure}

In a comprehensive review of all scenarios, it is evident that the majority of models fail to predict the correct OOD behavior initially, but demonstrate a robust ability to adapt to the new system dynamics with fine-tuning, which answers RQ 2 and 3.
However, the TCN-FAE model stands as an exception, often failing to show improvement within the given epoch limit. 
It is also the sole model that exhibits a performance decline in one scenario after fine-tuning for 50 epochs.
A common trend observed across the tested models is an initial degradation in performance upon fine-tuning all model weights, as illustrated in Figure \ref{fig:ftune_q1+v3}. 
Nevertheless, for most models, this loss quickly reverses, while the TCN-FAE often stagnates, hitting a performance plateau. 
Despite its ability to produce accurate forecasts on the 'standard' scenario, the TCN-FAE does not seem optimally designed for a transfer learning scenario, as highlighted by this experiment.
While further fine-tuning could potentially enhance the model's performance, it does not appear to leverage the system knowledge to the same degree as the other models. 
This could potentially be attributed to the bottleneck in the latent dimension or the autoencoder objective, which is learned concurrently with the predictions. 
Further investigations into these potential factors might provide more insight into the limitations observed in the TCN-FAE model's performance across various scenarios.

\subsection{Experiment 3: Data Augmentation}

In this experiment, we aim to assess how effectively the models adapt to data augmentation techniques. 
The results are presented in Table \ref{table:expt3}. 
As anticipated, both data augmentation techniques marginally reduce the performance on the 'standard' scenario on which the models were initially trained. 
Given that the principal motivation for employing data augmentation is to enhance model robustness, we examined performance in a scenario involving a perturbation (Scenario 1) and another reflecting OOD system behavior (Scenario 5).

\begin{table*}[h]
\centering
\begin{tabular}{lccccccccc}
\toprule
\multirow{2}{*}{\textbf{Model}} & \multicolumn{3}{c}{\textbf{Standard}} & \multicolumn{3}{c}{\textbf{Scenario 1}} & \multicolumn{3}{c}{\textbf{Scenario 5}} \\
\cmidrule(lr){2-4} \cmidrule(lr){5-7} \cmidrule(lr){8-10} 
                                &\textbf{None} & \textbf{Noise} & \textbf{Warp} &\textbf{None} & \textbf{Noise} & \textbf{Warp} &\textbf{None} & \textbf{Noise} & \textbf{Warp} \\
\midrule
MLP                             & 0.32          & 0.34      & 0.47          & \textbf{0.98} &\textbf{0.95}  & \textbf{1.33} & 6.01  &  6.31     & 3.98\\
GRU                             & 0.33          & 0.40      & 0.49          &  2.97         & 2.98          & 3.59          & 4.79  & 5.64       &\textbf{2.03}\\
GRU-AR                          &\textbf{0.28}  &\textbf{0.30} &\textbf{0.46}& 2.05          & 1.53          &1.88  & \textbf{2.56}  &\textbf{3.10}     &2.48\\
TCN                             &\underline{0.38}& 0.39      & 0.64         & 5.65          &2.97           & 3.48         & 5.18 &  5.08       &3.39\\
TCN-FAE                          & 0.34          & 0.43     &0.51            &4.02           & 4.37          &2.83           &\underline{6.47}& \underline{6.43}& 2.10\\
Transformer                     &\underline{0.38}&0.54   & \underline{0.74}&\underline{15.55}& \underline{9.40}& 11.58          &4.13  & 5.48&2.92\\
Transf.-CE                      &0.37           &\underline{0.67}& 0.64     & 8.78          & 7.66          & \underline{12.06}          & 4.97  &5.26&\underline{3.99}\\
\bottomrule
\end{tabular}
\caption{Performance comparison of different models fine-tuned on data augmented with various techniques. 
This table showcases the MSE values derived from models initially trained on the 'standard' scenario and subsequently fine-tuned using the respective data augmentation techniques (added noise or time warping). 
Lower MSE values indicate better performance. 
For each scenario and data augmentation technique, the best performing model is highlighted in bold, while the least performing model is underlined. 
Scenarios encompass the 'standard' scenario, 'sensor faults' (Scenario 1), and 'independent process phase merging' (Scenario 5).}
\label{table:expt3}
\end{table*}

The introduction of noise to the time series appears beneficial when managing perturbations such as faulty sensors, as most models exhibit improved performance on Scenario 1. 
However, this trend is not universally observed, as a few models display slightly diminished performance. 
Given that the performance on scenarios with OOD behavior and the 'standard' scenarios is compromised for most models, the indiscriminate application of this data augmentation technique for all model architectures is not recommended.

The impact of time warping diverges from the other data augmentation technique. 
While it generally enhances performance in the face of perturbations, the inclusion of noise proves more advantageous in Scenario 1. 
Nevertheless, time warping substantially improves performance when confronted with OOD behavior in Scenario 5 for all models. 
However, this performance enhancement comes with a trade-off: a more pronounced performance decline on the 'standard' scenario compared to the other data augmentation technique.

To answer RQ 4, we analyzed the individual models responses to data augmentation techniques. 
The MLP, demonstrating good performance across most scenarios, doesn't derive as much benefit from data augmentation as other models, suggesting cautious application. 
The same applies to the recurrent models, with the exception of the GRU with an auto-regressive forecasting method, which significantly enhances its predictions when dealing with sensor faults. 
Convolutional methods and models utilizing attention mechanisms, initially underperforming, mostly see improved robustness from data augmentation techniques, particularly the TCN variants. 
The TCN-FAE model responds exceptionally well to time warping. 
This trend warrants further investigation, necessitating a more extensive study.

\section{Conclusion and Future Work}
\label{conclusion}

This paper aimed to assess the robustness and generalization performance of various DL architectures applied to CPS data. 
The evaluated simulation indicated that recurrent architectures demonstrated robustness against perturbations, while convolutional methods exhibited the ability to generalize to certain OOD scenarios. 
Notably, MLP exhibited impressive performance in both areas. 
Conversely, Transformer models showed significant susceptibility to perturbations and struggled with generalization. 
However, this observation must be tempered by the fact that hardware limitations restricted the size and complexity of the Transformer models we could train in this study.
Most models demonstrated a successful transfer of knowledge to related tasks after fine-tuning, particularly the recurrent architectures and the MLP, with a noted exception being a specific TCN variant.

The findings of this study carry important implications for the application of DL methods. 
Firstly, practitioners must approach the application of DL models in CPS with caution, as performance disparities only became apparent when subjecting the models to specific test scenarios. 
Secondly, our study indicates that even relatively simple models, like MLPs, can perform surprisingly well with time series data. 
Lastly, despite most models initially struggling with generalization, fine-tuning proved to be effective. 
This suggests that fine-tuning models from related CPS tasks might be more cost-effective than training new models from scratch.

Our study also has several limitations. 
The underlying three-tank system is relatively simple and predictable. 
Future work could enhance the realism of the CPS simulation by incorporating control loops and discrete sensor values. 
Additionally, while the models tested provide a broad overview of DL architectures, newer variants that address existing limitations should be assessed to present a more accurate representation of the current state-of-the-art. 
Finally, alleviating hardware constraints would enable testing of more complex models, like large Transformers, thereby harnessing their full potential. 
Future research should aim to address these limitations and investigate more robust and generalizable DL architectures, such as causal or modular models.

Through this study, we have gained valuable insights into the application of DL models on CPS data. 
Our findings pave the way for more robust and reliable real-world CPS applications, thus contributing to the continual advancement of this domain.

\section*{Acknowledgments}
This research paper is funded by dtec.bw – Digitalization and Technology Research Center of the Bundeswehr.

\bibliographystyle{named}
\bibliography{ijcai23}

\begin{thebibliography}{}

\bibitem[\protect\citeauthoryear{Bansal \bgroup \em et al.\egroup
  }{2021}]{bansalSystematicGeneralizationNeural2021}
Hritik Bansal, Gantavya Bhatt, Pankaj Malhotra, and Prathosh AP.
\newblock Systematic {{Generalization}} in {{Neural Networks-based Multivariate
  Time Series Forecasting Models}}.
\newblock In {\em 2021 {{International Joint Conference}} on {{Neural
  Networks}} ({{IJCNN}})}, pages 1--8, July 2021.

\bibitem[\protect\citeauthoryear{Bayer \bgroup \em et al.\egroup
  }{2022}]{bayerSurveyDataAugmentation2022}
Markus Bayer, Marc-Andr{\'e} Kaufhold, and Christian Reuter.
\newblock A {{Survey}} on {{Data Augmentation}} for {{Text Classification}}.
\newblock {\em ACM Computing Surveys}, 55(7):146:1--146:39, December 2022.

\bibitem[\protect\citeauthoryear{Bengio \bgroup \em et al.\egroup
  }{2013}]{Bengio2013-yv}
Yoshua Bengio, Aaron Courville, and Pascal Vincent.
\newblock Representation learning: a review and new perspectives.
\newblock {\em IEEE Trans. Pattern Anal. Mach. Intell.}, 35(8):1798--1828,
  August 2013.

\bibitem[\protect\citeauthoryear{Bengio \bgroup \em et al.\egroup
  }{2021}]{bengioDeepLearningAI2021}
Yoshua Bengio, Yann Lecun, and Geoffrey Hinton.
\newblock Deep learning for {{AI}}.
\newblock {\em Communications of the ACM}, 64(7):58--65, June 2021.

\bibitem[\protect\citeauthoryear{Bottou}{2012}]{Bottou2012-sv}
L{\'e}on Bottou.
\newblock Stochastic gradient descent tricks.
\newblock In Gr{\'e}goire Montavon, Genevi{\`e}ve~B Orr, and Klaus-Robert
  M{\"u}ller, editors, {\em Neural Networks: Tricks of the Trade: Second
  Edition}, pages 421--436. Springer Berlin Heidelberg, Berlin, Heidelberg,
  2012.

\bibitem[\protect\citeauthoryear{Brown \bgroup \em et al.\egroup
  }{2020}]{brownLanguageModelsAre2020}
Tom Brown, Benjamin Mann, Nick Ryder, Melanie Subbiah, Jared~D Kaplan, Prafulla
  Dhariwal, Arvind Neelakantan, Pranav Shyam, Girish Sastry, Amanda Askell,
  Sandhini Agarwal, Ariel {Herbert-Voss}, Gretchen Krueger, Tom Henighan, Rewon
  Child, Aditya Ramesh, Daniel Ziegler, Jeffrey Wu, Clemens Winter, Chris
  Hesse, Mark Chen, Eric Sigler, Mateusz Litwin, Scott Gray, Benjamin Chess,
  Jack Clark, Christopher Berner, Sam McCandlish, Alec Radford, Ilya Sutskever,
  and Dario Amodei.
\newblock Language {{Models}} are {{Few-Shot Learners}}.
\newblock In {\em Advances in {{Neural Information Processing Systems}}},
  volume~33, pages 1877--1901. {Curran Associates, Inc.}, 2020.

\bibitem[\protect\citeauthoryear{Buzhinsky \bgroup \em et al.\egroup
  }{2021}]{buzhinskyMetricsMethodsRobustness2021}
Igor Buzhinsky, Arseny Nerinovsky, and Stavros Tripakis.
\newblock Metrics and methods for robustness evaluation of neural networks with
  generative models.
\newblock {\em Machine Learning}, July 2021.

\bibitem[\protect\citeauthoryear{Chen \bgroup \em et al.\egroup
  }{2022}]{chenDataDrivenApproachAssessing2022}
Chuang Chen, Ningyun Lu, Bin Jiang, and Yin Xing.
\newblock A {{Data-Driven Approach}} for {{Assessing Aero-Engine Health
  Status}}.
\newblock {\em IFAC-PapersOnLine}, 55(6):737--742, January 2022.

\bibitem[\protect\citeauthoryear{Cho \bgroup \em et al.\egroup
  }{2014}]{choLearningPhraseRepresentations2014}
Kyunghyun Cho, Bart {van Merri{\"e}nboer}, Caglar Gulcehre, Dzmitry Bahdanau,
  Fethi Bougares, Holger Schwenk, and Yoshua Bengio.
\newblock Learning {{Phrase Representations}} using {{RNN
  Encoder}}\textendash{{Decoder}} for {{Statistical Machine Translation}}.
\newblock In {\em Proceedings of the 2014 {{Conference}} on {{Empirical
  Methods}} in {{Natural Language Processing}} ({{EMNLP}})}, pages 1724--1734,
  {Doha, Qatar}, October 2014. {Association for Computational Linguistics}.

\bibitem[\protect\citeauthoryear{{de Beaulieu} \bgroup \em et al.\egroup
  }{2022}]{debeaulieuUnsupervisedRemainingUseful2022}
Martin~Herv{\'e} {de Beaulieu}, Mayank~Shekhar Jha, Hugues Garnier, and Farid
  Cerbah.
\newblock Unsupervised {{Remaining Useful Life Prediction}} through {{Long
  Range Health Index Estimation}} based on {{Encoders-Decoders}}.
\newblock {\em IFAC-PapersOnLine}, 55(6):718--723, January 2022.

\bibitem[\protect\citeauthoryear{Downs and
  Vogel}{1993}]{downsPlantwideIndustrialProcess1993}
J.~J. Downs and E.~F. Vogel.
\newblock A plant-wide industrial process control problem.
\newblock {\em Computers \& Chemical Engineering}, 17(3):245--255, March 1993.

\bibitem[\protect\citeauthoryear{Geneva and
  Zabaras}{2022}]{genevaTransformersModelingPhysical2022}
Nicholas Geneva and Nicholas Zabaras.
\newblock Transformers for modeling physical systems.
\newblock {\em Neural Networks}, 146:272--289, February 2022.

\bibitem[\protect\citeauthoryear{Goyal \bgroup \em et al.\egroup
  }{2021}]{goyalRecurrentIndependentMechanisms2021}
Anirudh Goyal, Alex Lamb, Jordan Hoffmann, Shagun Sodhani, Sergey Levine,
  Yoshua Bengio, and Bernhard Sch{\"o}lkopf.
\newblock Recurrent {{Independent Mechanisms}}.
\newblock In {\em 9th {{International Conference}} on {{Learning
  Representations}}, {{ICLR}} 2021, {{Virtual Event}}, {{Austria}}, {{May}}
  3-7, 2021}. {OpenReview.net}, 2021.

\bibitem[\protect\citeauthoryear{Goyal \bgroup \em et al.\egroup
  }{2022}]{goyalCoordinationNeuralModules2022}
Anirudh Goyal, Aniket~Rajiv Didolkar, Alex Lamb, Kartikeya Badola, Nan~Rosemary
  Ke, Nasim Rahaman, Jonathan Binas, Charles Blundell, Michael~Curtis Mozer,
  and Yoshua Bengio.
\newblock Coordination {{Among Neural Modules Through}} a {{Shared Global
  Workspace}}.
\newblock In {\em International {{Conference}} on {{Learning
  Representations}}}, January 2022.

\bibitem[\protect\citeauthoryear{Hochreiter and
  Schmidhuber}{1997}]{hochreiterLongShortTermMemory1997}
Sepp Hochreiter and J{\"u}rgen Schmidhuber.
\newblock Long {{Short-Term Memory}} | {{Neural Computation}}.
\newblock https://dl.acm.org/doi/10.1162/neco.1997.9.8.1735, 1997.

\bibitem[\protect\citeauthoryear{Khan \bgroup \em et al.\egroup
  }{2021}]{khanRobustnessAIbasedPrognostic2021}
Samir Khan, Seiji Tsutsumi, Takehisa Yairi, and Shinichi Nakasuka.
\newblock Robustness of {{AI-based}} prognostic and systems health management.
\newblock {\em Annual Reviews in Control}, 51:130--152, January 2021.

\bibitem[\protect\citeauthoryear{Kingma and
  Ba}{2015}]{DBLP:journals/corr/KingmaB14}
Diederik~P. Kingma and Jimmy Ba.
\newblock Adam: {{A}} method for stochastic optimization.
\newblock In Yoshua Bengio and Yann LeCun, editors, {\em 3rd International
  Conference on Learning Representations, {{ICLR}} 2015, San Diego, {{CA}},
  {{USA}}, May 7-9, 2015, Conference Track Proceedings}, 2015.

\bibitem[\protect\citeauthoryear{K{\"o}cher \bgroup \em et al.\egroup
  }{2022}]{kocherResearchAgendaAI2022}
Aljosha K{\"o}cher, Ren{\'e} Heesch, Niklas Widulle, Anna Nordhausen, Julian
  Putzke, Alexander Windmann, and Oliver Niggemann.
\newblock A {{Research Agenda}} for {{AI Planning}} in the {{Field}} of
  {{Flexible Production Systems}}.
\newblock In {\em 2022 {{IEEE}} 5th {{International Conference}} on
  {{Industrial Cyber-Physical Systems}} ({{ICPS}})}, pages 1--8, May 2022.

\bibitem[\protect\citeauthoryear{Koutroulis \bgroup \em et al.\egroup
  }{2022}]{koutroulisConstructingRobustHealth2022}
Georgios Koutroulis, Belgin Mutlu, and Roman Kern.
\newblock Constructing robust health indicators from complex engineered systems
  via anticausal learning.
\newblock {\em Engineering Applications of Artificial Intelligence},
  113:104926, August 2022.

\bibitem[\protect\citeauthoryear{Lea \bgroup \em et al.\egroup
  }{2017}]{leaTemporalConvolutionalNetworks2017}
Colin Lea, Michael~D. Flynn, Rene Vidal, Austin Reiter, and Gregory~D. Hager.
\newblock Temporal {{Convolutional Networks}} for {{Action Segmentation}} and
  {{Detection}}.
\newblock In {\em Proceedings of the {{IEEE Conference}} on {{Computer Vision}}
  and {{Pattern Recognition}}}, pages 156--165, 2017.

\bibitem[\protect\citeauthoryear{Lee \bgroup \em et al.\egroup
  }{2014}]{leePrognosticsHealthManagement2014}
Jay Lee, Fangji Wu, Wenyu Zhao, Masoud Ghaffari, Linxia Liao, and David Siegel.
\newblock Prognostics and health management design for rotary machinery
  systems\textemdash{{Reviews}}, methodology and applications.
\newblock {\em Mechanical Systems and Signal Processing}, 42(1):314--334,
  January 2014.

\bibitem[\protect\citeauthoryear{Li \bgroup \em et al.\egroup
  }{2019}]{liEnhancingLocalityBreaking2019}
Shiyang Li, Xiaoyong Jin, Yao Xuan, Xiyou Zhou, Wenhu Chen, Yu-Xiang Wang, and
  Xifeng Yan.
\newblock Enhancing the {{Locality}} and {{Breaking}} the {{Memory Bottleneck}}
  of {{Transformer}} on {{Time Series Forecasting}}.
\newblock In {\em Advances in {{Neural Information Processing Systems}}},
  volume~32. {Curran Associates, Inc.}, 2019.

\bibitem[\protect\citeauthoryear{Li \bgroup \em et al.\egroup
  }{2020}]{liAdversarialAttacksDefenses2020}
Jiao Li, Yang Liu, Tao Chen, Zhen Xiao, Zhenjiang Li, and Jianping Wang.
\newblock Adversarial {{Attacks}} and {{Defenses}} on
  {{Cyber}}\textendash{{Physical Systems}}: {{A Survey}}.
\newblock {\em IEEE Internet of Things Journal}, 7(6):5103--5115, June 2020.

\bibitem[\protect\citeauthoryear{Lippe \bgroup \em et al.\egroup
  }{2022a}]{lippeICITRISCausalRepresentation2022}
Phillip Lippe, Sara Magliacane, Sindy L{\"o}we, Yuki~M. Asano, Taco Cohen, and
  Efstratios Gavves.
\newblock {{iCITRIS}}: {{Causal Representation Learning}} for {{Instantaneous
  Temporal Effects}}, June 2022.

\bibitem[\protect\citeauthoryear{Lippe \bgroup \em et al.\egroup
  }{2022b}]{lippeCITRISCausalIdentifiability2022}
Phillip Lippe, Sara Magliacane, Sindy L{\"o}we, Yuki~M. Asano, Taco Cohen, and
  Stratis Gavves.
\newblock {{CITRIS}}: {{Causal Identifiability}} from {{Temporal Intervened
  Sequences}}.
\newblock In {\em Proceedings of the 39th {{International Conference}} on
  {{Machine Learning}}}, pages 13557--13603. {PMLR}, June 2022.

\bibitem[\protect\citeauthoryear{Luo \bgroup \em et al.\egroup
  }{2021}]{luoDeepLearningbasedAnomaly2021}
Yuan Luo, Ya~Xiao, Long Cheng, Guojun Peng, and Danfeng~(Daphne) Yao.
\newblock Deep {{Learning-based Anomaly Detection}} in {{Cyber-physical
  Systems}}: {{Progress}} and {{Opportunities}}.
\newblock {\em ACM Computing Surveys}, 54(5):106:1--106:36, May 2021.

\bibitem[\protect\citeauthoryear{Meng \bgroup \em et al.\egroup
  }{2020}]{mengTimeConvolutionalNetwork2020}
Chao Meng, Xue~Song Jiang, Xiu~Mei Wei, and Tao Wei.
\newblock A {{Time Convolutional Network Based Outlier Detection}} for
  {{Multidimensional Time Series}} in {{Cyber-Physical-Social Systems}}.
\newblock {\em IEEE Access}, 8:74933--74942, 2020.

\bibitem[\protect\citeauthoryear{Niggemann \bgroup \em et al.\egroup
  }{2023}]{niggemannMachineLearningCyberPhysical2023}
Oliver Niggemann, Bernd Zimmering, Henrik Steude, Jan~Lukas Augustin, Alexander
  Windmann, and Samim Multaheb.
\newblock Machine {{Learning}} for {{Cyber-Physical Systems}}.
\newblock In Birgit {Vogel-Heuser} and Manuel Wimmer, editors, {\em Digital
  {{Transformation}}: {{Core Technologies}} and {{Emerging Topics}} from a
  {{Computer Science Perspective}}}, pages 415--446. {Springer}, {Berlin,
  Heidelberg}, 2023.

\bibitem[\protect\citeauthoryear{Sch{\"o}lkopf \bgroup \em et al.\egroup
  }{2021}]{scholkopfCausalRepresentationLearning2021}
Bernhard Sch{\"o}lkopf, Francesco Locatello, Stefan Bauer, Nan~Rosemary Ke, Nal
  Kalchbrenner, Anirudh Goyal, and Yoshua Bengio.
\newblock Toward {{Causal Representation Learning}}.
\newblock {\em Proceedings of the IEEE}, 109(5):612--634, May 2021.

\bibitem[\protect\citeauthoryear{Shorten and
  Khoshgoftaar}{2019}]{shortenSurveyImageData2019}
Connor Shorten and Taghi~M. Khoshgoftaar.
\newblock A survey on {{Image Data Augmentation}} for {{Deep Learning}}.
\newblock {\em Journal of Big Data}, 6(1):60, July 2019.

\bibitem[\protect\citeauthoryear{Tsipras \bgroup \em et al.\egroup
  }{2019}]{tsiprasRobustnessMayBe2019}
Dimitris Tsipras, Shibani Santurkar, Logan Engstrom, Alexander Turner, and
  Aleksander Madry.
\newblock Robustness {{May Be}} at {{Odds}} with {{Accuracy}}.
\newblock In {\em International {{Conference}} on {{Learning
  Representations}}}, 2019.

\bibitem[\protect\citeauthoryear{Vaswani \bgroup \em et al.\egroup
  }{2017}]{vaswaniAttentionAllYou2017}
Ashish Vaswani, Noam Shazeer, Niki Parmar, Jakob Uszkoreit, Llion Jones,
  Aidan~N Gomez, {\L}ukasz Kaiser, and Illia Polosukhin.
\newblock Attention is {{All}} you {{Need}}.
\newblock In {\em Advances in {{Neural Information Processing Systems}}},
  volume~30. {Curran Associates, Inc.}, 2017.

\bibitem[\protect\citeauthoryear{Wen \bgroup \em et al.\egroup
  }{2021}]{wenTimeSeriesData2021}
Qingsong Wen, Liang Sun, Fan Yang, Xiaomin Song, Jingkun Gao, Xue Wang, and
  Huan Xu.
\newblock Time {{Series Data Augmentation}} for {{Deep Learning}}: {{A
  Survey}}.
\newblock In {\em Twenty-{{Ninth International Joint Conference}} on
  {{Artificial Intelligence}}}, volume~5, pages 4653--4660, August 2021.

\bibitem[\protect\citeauthoryear{Yin \bgroup \em et al.\egroup
  }{2021}]{yinLeveragingDataAugmentation2021}
Yuyu Yin, Haoran Xu, Tingting Liang, Manman Chen, Honghao Gao, and Antonella
  Longo.
\newblock Leveraging {{Data Augmentation}} for {{Service QoS Prediction}} in
  {{Cyber-physical Systems}}.
\newblock {\em ACM Transactions on Internet Technology}, 21(2):35:1--35:25,
  March 2021.

\bibitem[\protect\citeauthoryear{Zerveas \bgroup \em et al.\egroup
  }{2021}]{zerveasTransformerbasedFrameworkMultivariate2021}
George Zerveas, Srideepika Jayaraman, Dhaval Patel, Anuradha Bhamidipaty, and
  Carsten Eickhoff.
\newblock A {{Transformer-based Framework}} for {{Multivariate Time Series
  Representation Learning}}.
\newblock In {\em Proceedings of the 27th {{ACM SIGKDD Conference}} on
  {{Knowledge Discovery}} \& {{Data Mining}}}, {{KDD}} '21, pages 2114--2124,
  {New York, NY, USA}, August 2021. {Association for Computing Machinery}.

\bibitem[\protect\citeauthoryear{Zhao \bgroup \em et al.\egroup
  }{2023}]{zhaoGraphDataAugmentation2023}
Tong Zhao, Wei Jin, Yozen Liu, Yingheng Wang, Gang Liu, Stephan G{\"u}nnemann,
  Neil Shah, and Meng Jiang.
\newblock Graph {{Data Augmentation}} for {{Graph Machine Learning}}: {{A
  Survey}}, January 2023.

\bibitem[\protect\citeauthoryear{Zimmering \bgroup \em et al.\egroup
  }{2021}]{zimmeringGeneratingArtificialSensor2021}
Bernd Zimmering, Oliver Niggemann, Constanze Hasterok, Erik Pfannstiel, Dario
  Ramming, and Julius Pfrommer.
\newblock Generating {{Artificial Sensor Data}} for the {{Comparison}} of
  {{Unsupervised Machine Learning Methods}}.
\newblock {\em Sensors}, 21(7):2397, January 2021.

\end{thebibliography}

\end{document}